\pdfoutput=1

\documentclass[11pt]{article}

\usepackage[preprint]{acl}

\usepackage{times}
\usepackage{latexsym}

\usepackage[T1]{fontenc}

\usepackage[utf8]{inputenc}

\usepackage{microtype}

\usepackage{inconsolata}

\usepackage{graphicx}

\usepackage{booktabs}
\usepackage{amsmath}
\usepackage{subcaption}
\usepackage{soul}
\setul{0.3ex}{0.2ex}
\setulcolor{red}
\usepackage{multirow}

\usepackage{siunitx}
\usepackage{etoolbox}
\usepackage[normalem]{ulem}
\robustify\bfseries
\robustify\uline

\usepackage{tcolorbox}
\tcbuselibrary{listings}
\newtcolorbox{promptbox}[1][]{
  colback=gray!5,    
  colframe=gray!50,  
  fonttitle=\bfseries,
  coltitle=black,
  title=#1,
  boxrule=0.5pt,
  arc=2mm,           
  outer arc=2mm,
  left=4pt,right=4pt,top=4pt,bottom=4pt,
  after skip=12pt,
  listing only,
  listing options={
    basicstyle=\ttfamily\small,
    breaklines=true,
    columns=fullflexible
  }
}

\title{Mind the Gap:\textvisiblespace A Closer Look at Tokenization for\\Multiple-Choice Question Answering with LLMs}

\author{
    Mario Sanz-Guerrero$^\nabla$\quad~ Minh Duc Bui$^\nabla$\quad~ Katharina von der Wense$^{\nabla\spadesuit}$ \\
    $^\nabla$Johannes Gutenberg University Mainz, Germany \\
    $^\spadesuit$University of Colorado Boulder, USA \\
    \texttt{\{\href{mailto:msanz@uni-mainz.de}{msanz}, \href{mailto:minhducbui@uni-mainz.de}{minhducbui}, \href{mailto:k.vonderwense@uni-mainz.de}{k.vonderwense}\}@uni-mainz.de}
}

\begin{document}
\maketitle
\begin{abstract}
When evaluating large language models (LLMs) with multiple-choice question answering (MCQA), it is common to end the prompt with the string ``\texttt{Answer:}'' to facilitate automated answer extraction via next-token probabilities. However, there is no consensus on how to tokenize the space following the colon, often overlooked as a trivial choice. In this paper, we uncover accuracy differences of up to 11\% due to this (seemingly irrelevant) tokenization variation as well as reshuffled model rankings, raising concerns about the reliability of LLM comparisons in prior work. Surprisingly, we are able to recommend one specific strategy -- tokenizing the space \emph{together} with the answer letter -- as we observe consistent and statistically significant performance improvements. Additionally, it improves model calibration, enhancing the reliability of the model's confidence estimates. Our findings underscore the importance of careful evaluation design and highlight the need for standardized, transparent evaluation protocols to ensure reliable and comparable results.
\end{abstract}

\section{Introduction}
Leaderboards for evaluating large language models (LLMs) often include multiple-choice question answering (MCQA) tasks: the model is shown a question together with several candidate answers and must pick the correct one. To make automatic answer extraction easier, a widely used convention is to end the prompt with the literal string ``\texttt{Answer:}'' and look at the next-token probabilities for the option labels (usually: \texttt{A}/\texttt{B}/\texttt{C}/\texttt{D}).
This seemingly trivial formatting decision immediately poses another: \textit{should there be a space after the colon in the prompt, or should the space be included as part of the answer option token?}

\begin{figure}
    \centering
    \includegraphics[width=\linewidth]{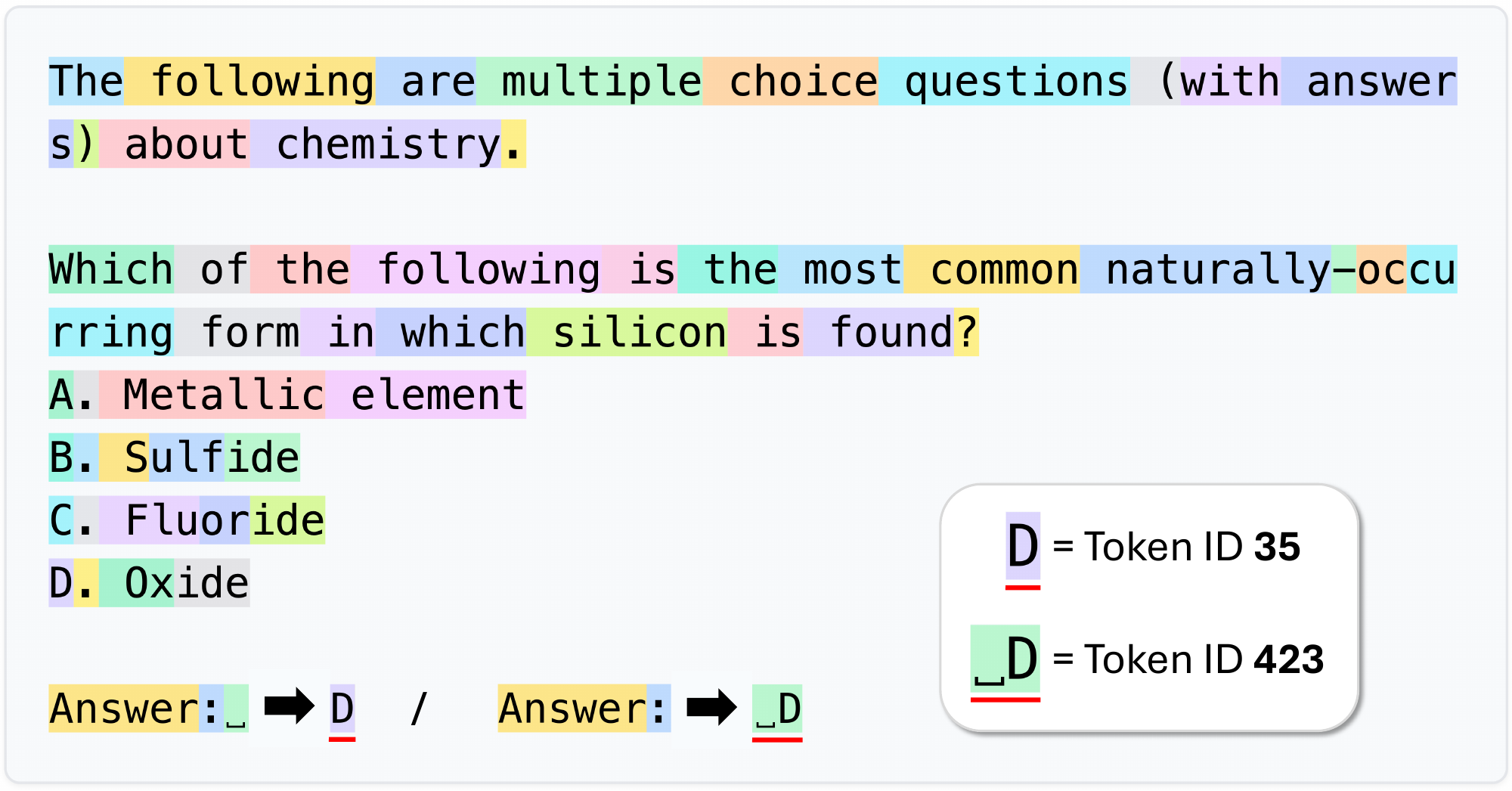}
    \caption{Illustration of a prompt tokenized with Llama~3.1. The final token representing the prediction depends on the tokenization of the space (``\texttt{\ul{D}}'', without space; or ``\texttt{\ul{\textvisiblespace D}}'', with space).}
    \label{fig:tokens_illustration}
\end{figure}

Recent studies have highlighted the significant performance variation that can arise from minor changes in prompt design \cite{zheng2024robust,pezeshkpour2024sensitivity}. However, little attention has been given to the role of tokenization, particularly the tokenization of the empty space character immediately preceding the answer label -- after the string ``\texttt{Answer:\textvisiblespace}'' (see Figure~\ref{fig:tokens_illustration}). More importantly, we note that \textit{practice is currently split}: some recent papers include the leading space in the prompt \cite{santurkar2023opinions,wang2024looktext,wang2024answerC}, while others omit it and tokenize it together with the letter \cite{zheng2024robust,hendrycks2021mmlu}, and no community-wide convention has emerged. Even widely used evaluation frameworks differ in their convention \cite{huggingface2023lighteval,eleuther2024lm-eval-harness}.

Surprisingly, we find significant differences in performance depending on the choice of the leading space tokenization. When the \textbf{leading space is tokenized together with the label letter, we observe consistent, statistically significant gains in both accuracy and calibration} across a wide range of LLMs and datasets.
This seemingly irrelevant tokenization choice alone shifts accuracy by as much as 11\% -- a larger effect than previously observed prompt formatting variations such as option order permutation.
Moreover, we find that \textbf{the choice of tokenization convention even alters model rankings}. When the space is tokenized before the letter, Llama 3.1 70B Instruct tops our leaderboard; when the space is tokenized together with the letter, Qwen 2.5 72B moves into first place.

Our experiments result in a clear recommendation: tokenize the space together with the letter, and observe model rankings exclusively for this configuration to ensure fair comparisons.
More generally, these findings underscore the need for unified evaluation frameworks and greater transparency, particularly for closed-source models, so LLM comparisons remain fair and meaningful.

\section{Related Work}

\paragraph{LLM Evaluation with MCQA}
Evaluating generative LLMs presents a significant challenge due to the open-ended nature of their outputs. Recent approaches have explored human evaluation and LLM-as-a-judge methods \cite{chiang2023llm-as-a-judge, chen2024humans}, but these techniques are highly subjective and unreliable. To address this, multiple-choice question answering has been widely adopted, as it enables automated, quantitative assessment of LLM capabilities.

There are multiple ways of automatically extracting an LLM's answer in MCQA tasks. Previous work have shown that better performance can be achieved by allowing the model to generate a free-form answer, followed by using a secondary LLM to extract the final choice \cite{wang2024answerC, lyu2024beyondprobabilities}. However, this approach is computationally expensive and can yield inconsistent results across different secondary models. Given that answer options are identified by letters (or labels), one of the most commonly used methodologies is to compute the model probabilities for the next token and get the highest label as the predicted choice \cite{hendrycks2021mmlu, santurkar2023opinions}.

\paragraph{Sensitivity to Prompt Details}
Recent studies have demonstrated that LLM performance in MCQA is highly sensitive to prompt details, often showing biases toward certain labels and answer order \cite{pezeshkpour2024sensitivity, zheng2024robust, alzahrani2024benchmarks}.
However, little attention has been given to the tokenization of the space character preceding the answer label, and there are discrepancies in the literature.
Some studies tokenize this space as an individual token \cite[e.g.,][]{santurkar2023opinions,wang2024looktext,wang2024answerC,pal2024geminimcqa}, while others tokenize it together with the answer label \cite[e.g.,][]{zheng2024robust,hendrycks2021mmlu}.

Complementarily to this body of research, we focus on a largely overlooked and \emph{apparently} irrelevant factor: the tokenization of the space character immediately preceding the answer label.

\section{Space or No Space?} \label{sec:motivation}
The literature has yet to converge on a single convention: even widely used evaluation frameworks such as Lighteval \cite{huggingface2023lighteval} from Hugging Face exhibit inconsistencies in how the leading space in the prompt is tokenized across different datasets. We begin by presenting the key arguments supporting each approach and identifying the prior works that have adopted them.

\paragraph{General Setting}
A MCQA prompt consists of a question and a set of answer choices, each associated with a distinct letter label as in Figure \ref{fig:tokens_illustration}. It finishes with the string ``\texttt{Answer:}'', and the LLM prediction is obtained as $\hat t = \arg\max_{t\in\{\mathrm{A},\mathrm{B},\dots\}} P(t\mid S)$, i.e., selecting the choice whose label token $t$ has the highest next-token conditional probability on the prompt $S$. This allows for an efficient and automated extraction of model answers for performance assessment.

\paragraph{Letter Token Without Space}
Given that the $t$ label tokens presented in the list of options in the prompt are tokenized as a single letter (without the leading space; see Figure \ref{fig:tokens_illustration}), it seems plausible to analyze the probability of the next token as a single letter as well (i.e., tokenizing the leading space as ``\texttt{Answer:\textvisiblespace}'', \emph{before} the actual token of the letter label). This tokenization represents the exact same token as the one in the corresponding option in the prompt
(more details in Appendix \ref{app:detailed_motivation}),
and there is a body of research that tokenizes this way \cite{santurkar2023opinions,wang2024looktext,wang2024answerC,pal2024geminimcqa}.

\paragraph{Letter Token With Space}
However, the previous is not the default tokenization: if we include the final answer letter in the prompt and tokenize it, the last token would be ``\texttt{\textvisiblespace$t$}''.
Thus, tokenizing the space \emph{together} with the letter also seems a reasonable approach.
This convention is also used in prior work \cite{zheng2024robust,hendrycks2021mmlu}.

\section{Experimental Setup}

The goal of our experiments is to analyze potential differences when tokenizing the option labels as single letters versus as letters preceded by a space (i.e., getting the model predictions from the ``\texttt{X}''\footnote{Where ``\texttt{X}'' is one of the option letters (\texttt{A}, \texttt{B}, \texttt{C}, \texttt{D}).} or ``\texttt{\textvisiblespace X}'' tokens, respectively). We look at it from two different perspectives: (1) performance, where we evaluate how accurate the model is in its predictions; and (2) calibration, where we assess how reliable the model predictions are.

\paragraph{Datasets}
Our main experiments are conducted on MMLU \cite{hendrycks2021mmlu}, one of the most widely used benchmarks for LLM evaluation \cite{openai2024gpt4, meta2024llama3}. MMLU contains multiple-choice questions from 57 different fields, providing a comprehensive set for interdisciplinary knowledge assessment. To ensure our findings are not specific to a single benchmark, we additionally evaluate on five other commonly used MCQA datasets (listed in Appendix \ref{app:datasets}).

\paragraph{Models}
We evaluate 15 LLMs from various families, sizes, and capabilities (listed in Appendix \ref{app:models}). All models are run with random sampling disabled (i.e., greedy decoding) for deterministic outputs and reproducibility.

\paragraph{Prompts}
To ensure our findings are robust and not limited to a single prompt template, we experiment with a variety of prompt formulations. These include zero-shot and few-shot settings, chain-of-thought (CoT) prompting, alternative formats for multiple-choice options, and prompts in different languages. Further details are provided in Appendix \ref{app:prompts}.

\paragraph{Evaluation}
For measuring performance, we report accuracy. As for calibration, we report the expected calibration error \cite[ECE;][]{Naeini2015ECE}, which measures the weighted average discrepancy between a model's prediction confidence and its actual accuracy across confidence bins. The formula of ECE is as follows:
\begin{equation*}
  \resizebox{\columnwidth}{!}{%
    $\displaystyle
      \mathrm{ECE}
      = \sum_{m=1}^M \frac{|B_m|}{N}
      \biggl|\,
        \underbrace{\frac{1}{|B_m|}\sum_{i \in B_m}\mathbf{1}\{\hat y_i = y_i\}}_{\mathrm{acc}(B_m)}
        \;-\;
        \underbrace{\frac{1}{|B_m|}\sum_{i \in B_m} p_i}_{\mathrm{conf}(B_m)}
      \,\biggr|
    $
  },
\end{equation*}
where $M$ is the number of confidence bins,
$N$ is the total number of instances,
$B_m$ is the set of instances whose predicted confidence falls into bin $m$,
$\hat y_i$ and $y_i$ are the predicted and true labels for instance $i$,
$p_i$ is the model's confidence for its predicted label on instance $i$,
$\mathrm{acc}(B_m)$ is the empirical accuracy in bin $m$,
and $\mathrm{conf}(B_m)$ is the average confidence in bin $m$. In our experiments, $M = 10$ bins (i.e., we have 10 bins, comprising 10\% accuracy each).

\paragraph{Statistical Significance}
To assess whether the choice of space tokenization strategy leads to statistically meaningful differences in performance, we conduct statistical tests comparing results from both setups.
For accuracy, we use McNemar's test \cite{McNemar1947} and, for calibration, we apply a paired bootstrap resampling test on the ECE (more details in Appendix \ref{app:stat_tests}).
In both cases, we consider differences significant when $p<0.05$.

\paragraph{Probability Extraction}
To obtain the model's predicted probabilities, we pass the prompt through the LLM and extract the next-token logits of the options letters. The logits are then converted into normalized probabilities, producing a probability distribution over all possible answers. This allows us to analyze not only the most likely answer but also how the model distributes probability mass (confidence) across all options, which is important for evaluating calibration.

\section{Results}

We evaluate our two tokenization schemes, which we denote: (1) \textbf{Letter token} (i.e., last line of the prompt is tokenized as \texttt{[``Answer'',``:'',``\textvisiblespace'',``\ul{X}'']}); and (2) \textbf{Space--Letter token} (i.e., last line of the prompt is tokenized as \texttt{[``Answer'',``:'',``\ul{\textvisiblespace X}'']}). We analyze the probability of the tokens \ul{underlined} in red.
Table~\ref{tab:results} shows the zero-shot results for all models on the MMLU dataset.

\begin{table}
    \centering
    \small
    \setlength{\tabcolsep}{5pt}
    \begin{tabular}{lSSSS}
        \toprule
        & \multicolumn{2}{c}{Accuracy (↑)} & \multicolumn{2}{c}{ECE (↓)} \\
        \cmidrule(lr){2-3} \cmidrule(lr){4-5}
        Model & ``\texttt{X}'' & ``\texttt{\textvisiblespace X}'' & ``\texttt{X}'' & ``\texttt{\textvisiblespace X}'' \\
        \midrule
        Llama 2 7B & 37.25 & \bfseries 38.88\textsuperscript{*} & 2.16 & \bfseries 1.15\textsuperscript{*} \\
        Llama 3.1 8B & 61.47 & \bfseries 63.93\textsuperscript{*} & 2.58 & \bfseries \phantom{0}\uline{0.50}\textsuperscript{*} \\
        Llama 3.1 8B Inst & 67.28 & \bfseries 68.73\textsuperscript{*} & 4.19 & \bfseries 3.77\textsuperscript{*} \\
        Llama 3.1 70B & 76.16 & \bfseries 76.64\textsuperscript{*} & 1.47 & \bfseries 1.16\textsuperscript{*} \\
        Llama 3.1 70B Inst & \uline{82.31}\phantom{\textsuperscript{*}} & \bfseries 82.60\textsuperscript{*} & \bfseries 3.91 & 4.87 \\
        Gemma 3 4B & 56.25 & \bfseries 57.95\textsuperscript{*} & 7.40 & \bfseries 1.74\textsuperscript{*} \\
        Gemma 3 4B Inst & 57.43 & \bfseries 57.77\textsuperscript{*} & \bfseries 20.34 & 20.36 \\
        Gemma 3 12B & 71.17 & \bfseries 71.31 & 2.18 & \bfseries 0.91\textsuperscript{*} \\
        Mistral 7B v0.3 & 60.17 & \bfseries 60.28 & 1.40 & \bfseries 0.51\textsuperscript{*} \\
        Mistral 7B Inst v0.3 & 59.70 & \bfseries 60.05\textsuperscript{*} & 12.83 & \bfseries 11.98\textsuperscript{*} \\
        Mistral Small 24B & 77.28 & \bfseries 77.66\textsuperscript{*} & \uline{0.79} & \bfseries 0.74 \\
        Qwen 2.5 7B & 69.38 & \bfseries 70.99\textsuperscript{*} & \bfseries 2.88 & 3.05 \\
        Qwen 2.5 72B & 81.93 & \bfseries \uline{83.24}\textsuperscript{*} & 1.10 & \bfseries 0.72 \\
        Qwen 3 8B & 72.82 & \bfseries 74.62\textsuperscript{*} & 2.95 & \bfseries 1.95\textsuperscript{*} \\
        GPT Neo 2.7B & 23.65 & \bfseries 24.39\textsuperscript{*} & 12.00 & \bfseries 4.05\textsuperscript{*} \\
        \bottomrule
    \end{tabular}
    \caption{Zero-shot performance on MMLU when tokenizing the answer letter as either a single letter (``\texttt{X}'') or as a space plus letter (``\texttt{\textvisiblespace X}''). \textsuperscript{*}~indicates a statistically significant improvement ($p < 0.05$). The top-performing model for each tokenization is underlined, and the top-performing tokenization strategy for each model is bolded.}
    \label{tab:results}
\end{table}

\paragraph{Performance}
While the place where the space preceding the answer letter is tokenized might seem completely irrelevant, we observe noteworthy \textbf{accuracy gains in all models when tokenizing the space \emph{within the same token} as the actual letter}. These improvements are statistically significant (except for Gemma 3 12B and Mistral 7B). This might seem counterintuitive, as these tokens are not the same as the ones in the options list of the prompt.

These performance differences have crucial practical implications: even by evaluating only a handful of models, we show noticeable changes in a hypothetical leaderboard. \textbf{By only changing where the space is tokenized, the top-performing model changes} (from Llama 3.1 70B Instruct to Qwen 2.5 72B). This indicates that such a subtle tweak could significantly alter LLM leaderboards.

\begin{figure}
    \centering
    \begin{subfigure}[b]{0.5\columnwidth}
        \centering
        \includegraphics[width=0.98\textwidth]{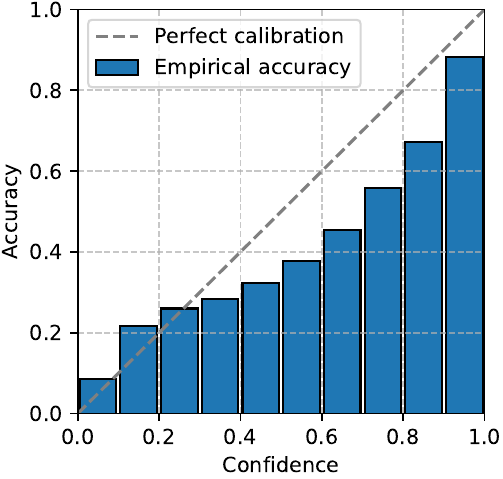}
        \caption{Letter token (``\texttt{X}'').}
        \label{subfig:reliability_before}
    \end{subfigure}%
    \begin{subfigure}[b]{0.5\columnwidth}
        \centering
        \includegraphics[width=0.98\textwidth]{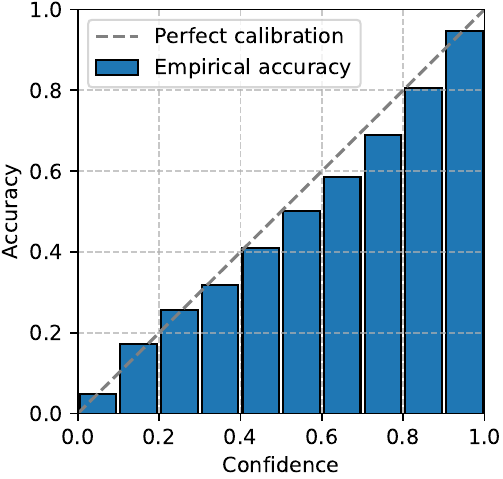}
        \caption{Space--Letter token (``\texttt{\textvisiblespace X}'').}
        \label{subfig:reliability_after}
    \end{subfigure}
    
    \caption{Reliability diagrams for Gemma 3 4B.}
    \label{fig:reliability}
\end{figure}

\paragraph{Calibration}
Additionally, the ECE is lower for the large majority of models when tokenizing the space with the letter, with many of the differences being significant under the paired bootstrap resampling test. We find \textbf{model answers being up to 4 times more reliable by only changing the tokenization of the leading space} (see Gemma 3 4B).
For illustrating the calibration of the models, Figure \ref{fig:reliability} shows the reliability diagrams for the Gemma 3 model. The main calibration gains come from the 30\% confidence bin onwards -- which, after tweaking the tokenization, become closer to the perfect calibration.
These improvements are crucial for enhanced performance, as even minor changes near the model's decision boundary can greatly affect predictions.

\subsection{Few-Shot and Chain-of-Thought Results}
\paragraph{Few-Shot}
In the few-shot scenario, we include 5 example questions and answers in the prompt before the target question, using the same tokenization for every answer as in the final (evaluated) token. This approach is widely used to help models better understand the task format and expected output \cite{openai2024gpt4,meta2024llama3}. The results in Table~\ref{tab:results_fs_cot} show that the accuracy and calibration improvements from space--letter tokenization persist in the few-shot setting, confirming that the impact of space tokenization is robust even when the model is provided with explicit demonstrations of the answer format.

\paragraph{Chain-of-Thought}
We further test the effect of space tokenization under CoT prompting, where the model is encouraged to reason step-by-step before providing its answer. Table~\ref{tab:results_fs_cot} shows that, while calibration still improves significantly, the absolute accuracy gains are not -- this is reasonable since, after the reasoning chain, extracting the answer label is more straightforward and thus less sensitive to the empty space tokenization.

\begin{table}
    \centering
    \small
    \setlength{\tabcolsep}{5pt}
    \begin{tabular}{lSSSS}
        \toprule
        & \multicolumn{2}{c}{Accuracy (↑)} & \multicolumn{2}{c}{ECE (↓)} \\
        \cmidrule(lr){2-3} \cmidrule(lr){4-5}
        Setting & ``\texttt{X}'' & ``\texttt{\textvisiblespace X}'' & ``\texttt{X}'' & ``\texttt{\textvisiblespace X}'' \\
        \midrule
        Zero-shot & 61.47 & \bfseries 63.93\textsuperscript{*} & 2.58 & \bfseries 0.50\textsuperscript{*} \\
        Few-shot & 63.90 & \bfseries 65.78\textsuperscript{*} & 2.24 & \bfseries 0.37\textsuperscript{*} \\
        Chain-of-Thought & 69.64 & \bfseries 70.11 & 6.36 & \bfseries 3.75\textsuperscript{*} \\
        \bottomrule
    \end{tabular}
    \caption{Performance of Llama 3.1 8B on MMLU under different prompt settings.}
    \label{tab:results_fs_cot}
\end{table}

\subsection{Robustness to Prompt Variations}\label{sec:prompt_variations}
Recent work has demonstrated that LLMs are highly sensitive to subtle changes in prompt phrasing and structure. To evaluate the robustness of our findings, we experiment with a range of prompt formulations (see Appendix~\ref{app:prompt_variations} for details on these perturbations). Table~\ref{tab:prompt_variations} shows that the impact of empty space tokenization is consistent across all prompt variations and in fact exceeds the effects of other prompt modifications such as changing option labels or their order.\footnote{Numeric labels yield identical results because ``\texttt{\textvisiblespace$n$}'' is tokenized as [``\texttt{\textvisiblespace}'', ``\texttt{$n$}''], so the final token is the same for both tokenization strategies.} Furthermore, Appendix~\ref{app:languages} presents results for MMLU in five different languages -- including some not natively supported by the models -- and our findings remain robust in all cases.

\begin{table}
    \centering
    \small
    \setlength{\tabcolsep}{4pt}
    \begin{tabular}{lcccc}
        \toprule
        & \multicolumn{2}{c}{Accuracy (↑)} & \multicolumn{2}{c}{ECE (↓)} \\
        \cmidrule(lr){2-3} \cmidrule(lr){4-5}
        Prompt Template & ``\texttt{X}'' & ``\texttt{\textvisiblespace X}'' & ``\texttt{X}'' & ``\texttt{\textvisiblespace X}'' \\
        \midrule
        Original & 61.47 & \bfseries 63.93\textsuperscript{*} & 2.58 & \bfseries 0.50\textsuperscript{*} \\
        \midrule
        Parentheses (``\texttt{\textvisiblespace(A)}'') & 62.07 & \bfseries 64.18\textsuperscript{*} & 1.92 & \bfseries 1.07\textsuperscript{*} \\
        Numbers (``\texttt{\textvisiblespace 1}'') & 62.21 & 62.21\phantom{\textsuperscript{*}} & 1.94 & 1.94\phantom{\textsuperscript{*}} \\
        Space in option list & 61.89 & \bfseries 63.25\textsuperscript{*} & 1.86 & \bfseries 0.74\textsuperscript{*} \\
        Choices before question & 44.52 & \bfseries 48.02\textsuperscript{*} & 6.82 & \bfseries 2.74\textsuperscript{*} \\
        Permutations (avg.) & 61.53 & \bfseries 63.37\textsuperscript{*} & 2.98 & \bfseries 0.56\textsuperscript{*} \\
        \bottomrule
    \end{tabular}
    \caption{Performance of Llama 3.1 8B on the MMLU benchmark with different prompt templates.}
    \label{tab:prompt_variations}
\end{table}

\subsection{Results on Other Datasets}
\begin{figure}
    \centering
    \includegraphics[width=\linewidth]{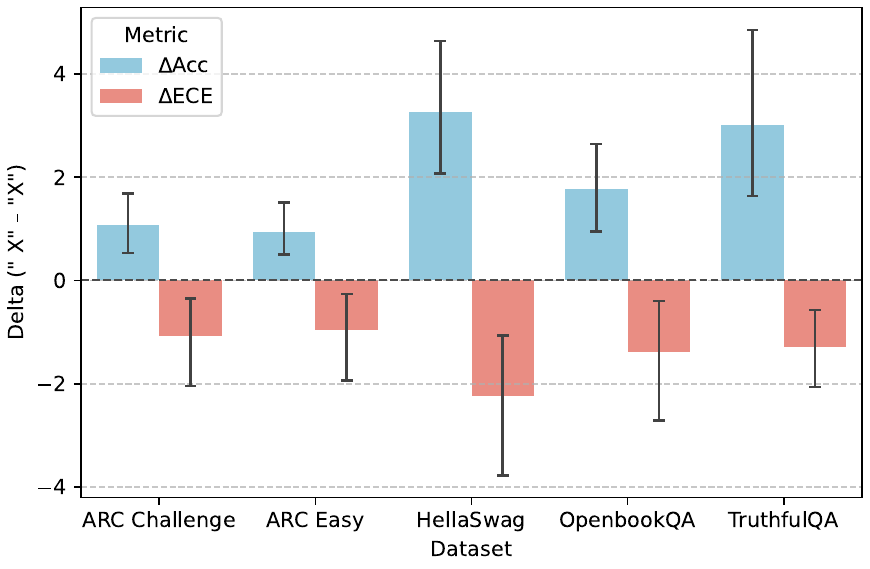}
    \caption{Mean accuracy improvement (left) and ECE reduction (right) from tokenizing the space with the answer letter, averaged across all models for each dataset. Error bars represent 95\% confidence intervals.}
    \label{fig:datasets_deltas}
\end{figure}

We further validate the generality of the effect by evaluating all models on five additional widely used MCQA datasets (ARC Challenge \cite{allenai2018arc}, ARC Easy \cite{allenai2018arc}, HellaSwag \cite{zellers2019hellaswag}, OpenbookQA \cite{mihaylov2018openbookqa}, and TruthfulQA \cite{lin2022truthfulqa}). Figure~\ref{fig:datasets_deltas} summarizes the average accuracy improvement and ECE reduction (delta aggregated across models) for each dataset, while the full per-model results are reported in Appendix~\ref{app:datasets_results}. The trends are consistent with the previous findings: tokenizing the space together with the answer letter systematically increases accuracy while (in most cases) lowering calibration error. Notably, even the largest model in our study (Qwen~2.5~72B) exhibits a very substantial accuracy gain of over 11\% on the HellaSwag dataset under the space--letter tokenization (``\texttt{\textvisiblespace X}''), underscoring that the effect is not confined to smaller or less capable models.

\section{Conclusion}
In this work, we uncover a subtle yet impactful detail in the MCQA evaluation of LLMs: the tokenization of the space preceding the answer letter.
Despite the lack of a standardized convention for this tokenization -- often dismissed as an irrelevant choice -- we show that it has significant implications for both model performance and reliability.
Our experiments reveal that tokenizing the space \emph{together with the option letter} leads to consistent improvements in accuracy and calibration, with performance gains reaching up to 11\%. More strikingly, this minor tokenization change is sufficient to alter the relative rankings of models on leaderboards, raising important concerns about the comparability of prior LLM evaluation results.
We encourage future work to consider these low-level details carefully to ensure fair and meaningful model comparisons.

\section*{Limitations}
Our evaluation focuses on open-weight models, as we require access to all next-token logits, which are not provided for proprietary, API-based models.
To allow for extensive experimentation with different models under our computational constraints, we use small- to medium-sized LLMs (up to 72B) and observe similar trends across all of them. Testing our findings with large-scale LLMs remains for future work.

\section*{Ethics Statement}
Our work highlights a noteworthy discrepancy in the current literature on LLM evaluation with MCQA and demonstrates significant performance improvements by tokenizing the empty space together with the subsequent answer letter token. However, these improvements do not fully eliminate inherent risks, and LLMs remain susceptible to errors. Therefore, we caution against relying solely on LLMs in critical settings, such as medical advice, without appropriate human oversight and domain-specific validation.

\section*{Acknowledgments}
This work was supported by the Carl Zeiss Foundation through the MAINCE and TOPML projects (grant numbers P2022-08-009 and P2021-02-014).

\bibliography{custom}

\appendix

\section{Space or No Space? Detailed Motivation of Both Approaches}
\label{app:detailed_motivation}

In Section \ref{sec:motivation} we discuss the two current approaches in the literature for the tokenization of the empty space preceding the answer letter. Here, we provide a more in-depth analysis of the rationale for both approaches.

\subsection{Letter Token Without Space: Token Similarity}
\begin{figure}
    \centering
    \includegraphics[width=\linewidth]{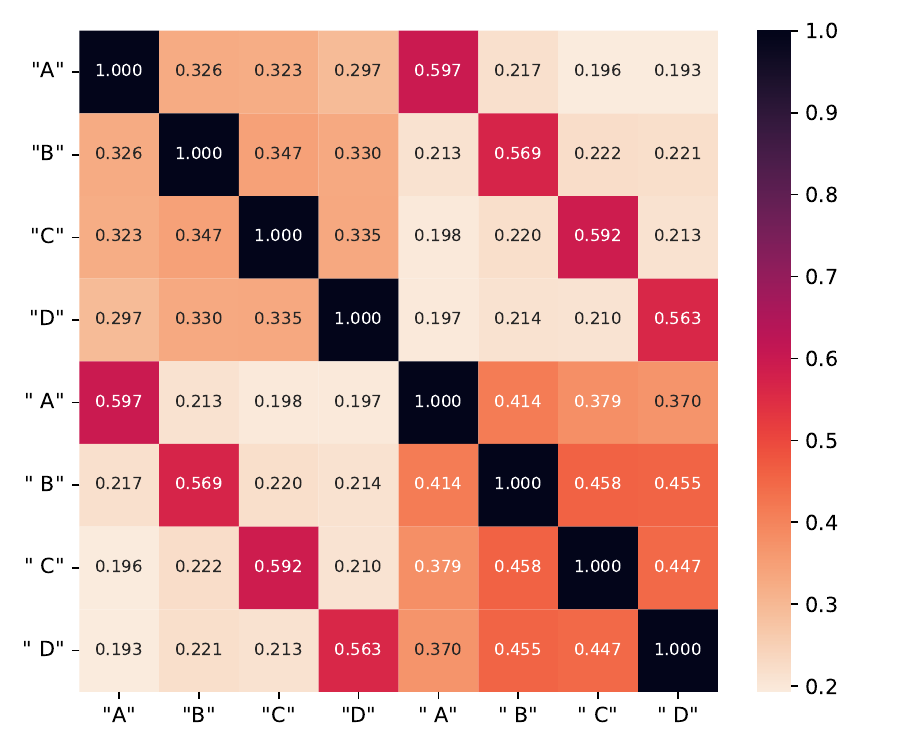}
    \caption{Cosine similarity of the Llama 3.1 token embeddings of the options, with and without space tokenization.}
    \label{fig:token_similarity_matrix}
\end{figure}

This approach involves tokenizing the space independently in the prompt template, and looking at the probability of the model generating the label tokens ``\texttt{$t$}'' (without leading space). Many studies tokenize this way \cite{santurkar2023opinions,wang2024answerC,wang2024looktext,pal2024geminimcqa}. The main potential reason is that the ``\texttt{$t$}'' tokens are exactly the same as the ones representing the corresponding options in the prompt (see Figure \ref{fig:tokens_illustration}).
A recent study highlights this exact match between the tokens in the options list and in the final answer as an important aspect \cite{gu2025olmes}.
To quantify the strength of this argument, Figure \ref{fig:token_similarity_matrix} shows the token embedding similarity of the option letters.

Consider a MCQA task whose correct choice is ``\texttt{X}''. When computing the model probabilities for each option (final token), the embeddings of the tokens ``\texttt{\textvisiblespace X}'' (correct option with space) and ``\texttt{\textvisiblespace Y}'' (incorrect option with space) compared to ``\texttt{X}'' (ground truth token) are much more similar among them ($\approx0.6$ vs.\ $\approx0.2$, respectively) than if we compare the tokens ``\texttt{X}'' (correct option without space) and ``\texttt{Y}'' (incorrect option without space) to ``\texttt{X}'' (1.0 vs.\ $\approx0.3$, respectively). Therefore, it seems reasonable to use the same token as in the list of options in the prompt (i.e., ``\texttt{$t$}'' tokens) since the embeddings of the letter labels are more easily distinguishable. In a situation where the model is in doubt between the two options, this could allow clearer decision boundaries among the choices.

\subsection{Letter Token With Space: Model's Default Tokenization}

On the other hand, this other approach involves tokenizing the leading space together with the option letter, extracting the model predictions from the probabilities of the ``\texttt{\textvisiblespace$t$}'' tokens. The rationale for this choice is that it represents the default tokenization of the model after including the letter in the prompt -- if we tokenize the string ``\texttt{Answer:~X}'', the last token would be ``\texttt{\textvisiblespace X}''. Thus, this tokenization aligns better with what the model would expect to see, so it seems a plausible approach as well.
This convention is also used in prior studies \cite{zheng2024robust,hendrycks2021mmlu}, and has been adopted for other tasks beyond MCQA, such as classification \cite{sanzguerrero2025cicl}.

\section{Detailed Experimental Setup}
\label{app:detailed_experimental_setup}

\subsection{Datasets}
\label{app:datasets}

Table \ref{tab:datasets_info} contains the list of datasets used in this study. All of them are evaluated using the default test set from Hugging Face\footnote{\url{https://huggingface.co/datasets}}, the size of which is specified in the table.

\begin{table}
    \centering
    \small
    \begin{tabular}{lr}
        \toprule
        Dataset & $|$Test$|$ \\
        \midrule
        MMLU \cite{hendrycks2021mmlu} & 14,042 \\
        AI2 ARC Easy \cite{allenai2018arc} & 2,365 \\
        AI2 ARC Challenge \cite{allenai2018arc} & 1,172 \\
        HellaSwag \cite{zellers2019hellaswag} & 10,003 \\
        OpenbookQA \cite{mihaylov2018openbookqa} & 500 \\
        TruthfulQA \cite{lin2022truthfulqa} & 817 \\
        \bottomrule
    \end{tabular}
    \caption{Datasets (and their sizes) used in this paper.}
    \label{tab:datasets_info}
\end{table}

\subsection{Models}
\label{app:models}

Table \ref{tab:models_info} contains the list of models used in this study. All of them are downloaded from the Hugging Face model hub\footnote{\url{https://huggingface.co/models}}, and their size is specified in the table.

\begin{table}
    \centering
    \small
    \begin{tabular}{l}
        \toprule
        Model  \\
        \midrule
        Llama 2 7B \cite{meta2023llama2}   \\
        Llama 3.1 8B \cite{meta2024llama3}   \\
        Llama 3.1 8B Instruct \cite{meta2024llama3}   \\
        Llama 3.1 70B \cite{meta2024llama3}  \\
        Llama 3.1 70B Instruct \cite{meta2024llama3}   \\
        Gemma 3 4B \cite{google2025gemma3}   \\
        Gemma 3 4B Instruct \cite{google2025gemma3}   \\
        Gemma 3 12B \cite{google2025gemma3}   \\
        Mistral 7B v0.3 \cite{jiang2023mistral7b}   \\
        Mistral 7B Instruct v0.3 \cite{jiang2023mistral7b}   \\
        Mistral Small 24B \cite{mistral2025Small3}   \\
        Qwen 2.5 7B \cite{qwen2025qwen25}   \\
        Qwen 2.5 72B \cite{qwen2025qwen25}   \\
        Qwen 3 8B \cite{yang2025qwen3}   \\
        GPT Neo 2.7B \cite{gpt-neo}   \\
        \bottomrule
    \end{tabular}
    \caption{LLMs evaluated in this paper.}
    \label{tab:models_info}
\end{table}

\subsection{Prompts}
\label{app:prompts}

\subsubsection{Main Prompt Templates}

Figures \ref{fig:base_prompt} and \ref{fig:instruct_prompt} show the prompts used for base and instruction-tuned models, respectively. The difference between the two prompts is the inclusion of special tokens (\{system token\}, \{user token\}, and \{assistant token\}) in the instruction-tuned models, which are model-specific and represent the expected usage of these models, aligning with conversational interactions. In the figures, the relevant space tokens are marked as \texttt{\textvisiblespace}. The last line of the prompt is the one that changes between the two tokenization strategies. The probability for each option is extracted from the next-token logits of the tokens after the arrow ($\rightarrow$), which are \ul{underlined} in red.

\begin{figure}[ht]
    \centering
    \begin{promptbox}[MCQA Main Prompt (Base models)]
        \ttfamily\small\raggedright
        ``The following are multiple choice questions (with answers).\\
        Question: \{question\}\\
        A. \{option A\}\\
        B. \{option B\}\\
        C. \{option C\}\\
        D. \{option D\}\\
        Answer:\textvisiblespace'' $\rightarrow$ ``\ul{X}'' // ``Answer:'' $\rightarrow$ ``\ul{\textvisiblespace X}''
    \end{promptbox}
    \caption{Prompt used for base models. The relevant space tokens are marked as \texttt{\textvisiblespace}. We analyze the probabilities of the tokens after the arrow ($\rightarrow$), which are \ul{underlined} in red. ``\texttt{X}'' denotes the option label (\texttt{A}/\texttt{B}/\texttt{C}/\texttt{D}).}
    \label{fig:base_prompt}
\end{figure}

\begin{figure}[ht]
    \centering
    \begin{promptbox}[MCQA Main Prompt (Instruction models)]
        \ttfamily\small\raggedright
        \{system token\}\\
        ``You are a helpful assistant for multiple-choice questions. Always answer strictly in the format ``Answer: X'', where X is the letter of the chosen answer (A, B, C, or D). Do not include any other text or explanation.''\\
        \{user token\}\\
        ``Question: \{question\}\\
        A. \{option A\}\\
        B. \{option B\}\\
        C. \{option C\}\\
        D. \{option D\}''\\
        \{assistant token\}\\
        ``Answer:\textvisiblespace'' $\rightarrow$ ``\ul{X}'' // ``Answer:'' $\rightarrow$ ``\ul{\textvisiblespace X}''
    \end{promptbox}
    \caption{Prompt used for instruction-tuned models. The relevant space tokens are marked as \texttt{\textvisiblespace}. We analyze the probabilities of the tokens after the arrow ($\rightarrow$), which are \ul{underlined} in red. ``\texttt{X}'' denotes the option label (\texttt{A}/\texttt{B}/\texttt{C}/\texttt{D}). The \{system token\}, \{user token\}, and \{assistant token\} are model-specific special tokens.}
    \label{fig:instruct_prompt}
\end{figure}

\subsubsection{Prompt Variations}
\label{app:prompt_variations}

Recent work has shown that LLMs are highly sensitive to subtle changes in prompt phrasing and structure \cite{pezeshkpour2024sensitivity, zheng2024robust, alzahrani2024benchmarks}. To ensure that our findings are robust to such variations, we experiment with a range of prompt formulations, as analyzed in Section \ref{sec:prompt_variations} (Table \ref{tab:prompt_variations}). Here, we provide the exact prompts used for each variation.

Figure \ref{fig:space_in_option_list} shows the prompt variation with a space before each option in the list. This modification ensures that the final answer token (``\texttt{\textvisiblespace X}'') matches the format of the options in the list (``\texttt{\textvisiblespace A}'', ``\texttt{\textvisiblespace B}'', etc.). Figure \ref{fig:parentheses} shows the prompt variation with parentheses around the option labels. Figure \ref{fig:numbers} shows the prompt variation with numeric option labels (1/2/3/4). Figure \ref{fig:choices_before_question} shows the prompt variation with the list of options before the question.

\begin{figure}[]
    \centering
    \begin{promptbox}[Prompt Variation: Space in Option List]
        \ttfamily\small\raggedright
        ``The following are multiple choice questions (with answers).\\
        Question: \{question\}\\
        \textvisiblespace A. \{option A\}\\
        \textvisiblespace B. \{option B\}\\
        \textvisiblespace C. \{option C\}\\
        \textvisiblespace D. \{option D\}\\
        Answer:\textvisiblespace'' $\rightarrow$ ``\ul{X}'' // ``Answer:'' $\rightarrow$ ``\ul{\textvisiblespace X}''
    \end{promptbox}
    \caption{Prompt variation with a space before each option in the list. The relevant space tokens are marked as \texttt{\textvisiblespace}. We analyze the probabilities of the tokens after the arrow ($\rightarrow$), which are \ul{underlined} in red. ``\texttt{X}'' denotes the option label (\texttt{A}/\texttt{B}/\texttt{C}/\texttt{D}).}
    \label{fig:space_in_option_list}
\end{figure}

\begin{figure}[]
    \centering
    \begin{promptbox}[Prompt Variation: Parentheses]
        \ttfamily\small\raggedright
        ``The following are multiple choice questions (with answers).\\
        Question: \{question\}\\
        (A) \{option A\}\\
        (B) \{option B\}\\
        (C) \{option C\}\\
        (D) \{option D\}\\
        Answer:\textvisiblespace'' $\rightarrow$ ``\ul{(X)}'' // ``Answer:'' $\rightarrow$ ``\ul{\textvisiblespace(X)}''
    \end{promptbox}
    \caption{Prompt variation with parentheses around the option labels. The relevant space tokens are marked as \texttt{\textvisiblespace}. We analyze the probabilities of the tokens after the arrow ($\rightarrow$), which are \ul{underlined} in red. ``\texttt{X}'' denotes the option label (\texttt{A}/\texttt{B}/\texttt{C}/\texttt{D}).}
    \label{fig:parentheses}
\end{figure}

\begin{figure}[]
    \centering
    \begin{promptbox}[Prompt Variation: Numbers]
        \ttfamily\small\raggedright
        ``The following are multiple choice questions (with answers).\\
        Question: \{question\}\\
        1. \{option A\}\\
        2. \{option B\}\\
        3. \{option C\}\\
        4. \{option D\}\\
        Answer:\textvisiblespace'' $\rightarrow$ ``\ul{n}'' // ``Answer:'' $\rightarrow$ ``\ul{\textvisiblespace n}''
    \end{promptbox}
    \caption{Prompt variation with numeric option labels. The relevant space tokens are marked as \texttt{\textvisiblespace}. We analyze the probabilities of the tokens after the arrow ($\rightarrow$), which are \ul{underlined} in red. ``\texttt{n}'' denotes the option label (1/2/3/4).}
    \label{fig:numbers}
\end{figure}

\begin{figure}[]
    \centering
    \begin{promptbox}[Prompt Variation: Choices Before Question]
        \ttfamily\small\raggedright
        ``The following are multiple choice questions (with answers).\\
        A. \{option A\}\\
        B. \{option B\}\\
        C. \{option C\}\\
        D. \{option D\}\\
        Question: \{question\}\\
        Answer:\textvisiblespace'' $\rightarrow$ ``\ul{X}'' // ``Answer:'' $\rightarrow$ ``\ul{\textvisiblespace X}''
    \end{promptbox}
    \caption{Prompt variation with the list of options before the question. The relevant space tokens are marked as \texttt{\textvisiblespace}. We analyze the probabilities of the tokens after the arrow ($\rightarrow$), which are \ul{underlined} in red. ``\texttt{X}'' denotes the option label (\texttt{A}/\texttt{B}/\texttt{C}/\texttt{D}).}
    \label{fig:choices_before_question}
\end{figure}

\subsection{Statistical Tests}
\label{app:stat_tests}

\paragraph{Accuracy: McNemar's Test}
For accuracy, we use McNemar's test \cite{McNemar1947}, which assesses whether the number of examples correctly answered only under the ``\texttt{\textvisiblespace X}'' tokenization is significantly greater than those only correct under the ``\texttt{X}'' tokenization.

\paragraph{Calibration: Paired Bootstrap Resampling Test}
For calibration, we apply a paired bootstrap resampling test on the ECE. We resample the evaluation examples with replacement and recompute the ECE difference for each bootstrap sample, estimating the probability that one tokenization strategy leads to significantly lower ECE.

\section{Additional Results}
\label{app:additional_results}

\subsection{Effect of Option Order Permutations}

Motivated by recent works showing that LLMs are sensitive to the order of options in MCQA prompts \cite{zheng2024robust,pezeshkpour2024sensitivity}, we experiment with 5 different random shufflings of the options in the prompt. These results are averaged in Table \ref{tab:prompt_variations} under ``Permutations (avg.)'', and the individual results for each shuffling are reported in Table \ref{tab:option_orders}. We observe that not only the improvements from tokenizing the space with the letter (compared across columns) are consistent, but also that this effect is larger than the differences caused by changing the order of the options (compared across rows).

\begin{table}[]
    \centering
    \small
    \begin{tabular}{lcccc}
        \toprule
        & \multicolumn{2}{c}{Accuracy (↑)} & \multicolumn{2}{c}{ECE (↓)} \\
        \cmidrule(lr){2-3} \cmidrule(lr){4-5}
        Option order & ``\texttt{X}'' & ``\texttt{\textvisiblespace X}'' & ``\texttt{X}'' & ``\texttt{\textvisiblespace X}'' \\
        \midrule
        Original & 61.47 & \bfseries 63.93\textsuperscript{*} & 2.58 & \bfseries 0.50\textsuperscript{*} \\
        \midrule
        Permutation 1 & 61.78 & \bfseries 63.86\textsuperscript{*} & 2.86 & \bfseries 0.64\textsuperscript{*} \\
        Permutation 2 & 61.43 & \bfseries 62.89\textsuperscript{*} & 3.29 & \bfseries 0.41\textsuperscript{*} \\
        Permutation 3 & 62.11 & \bfseries 63.95\textsuperscript{*} & 2.51 & \bfseries 0.61\textsuperscript{*} \\
        Permutation 4 & 61.28 & \bfseries 63.29\textsuperscript{*} & 2.93 & \bfseries 0.52\textsuperscript{*} \\
        Permutation 5 & 61.07 & \bfseries 62.85\textsuperscript{*} & 3.32 & \bfseries 0.61\textsuperscript{*} \\
        \bottomrule
    \end{tabular}
    \caption{Performance of Llama 3.1 8B on the MMLU benchmark with different random option orders.}
    \label{tab:option_orders}
\end{table}

\subsection{Performance across Languages}
\label{app:languages}

Table~\ref{tab:languages} demonstrates that the accuracy and calibration improvements from space--letter tokenization (``\texttt{\textvisiblespace X}'') are robust across multiple languages. For all tested languages, including Spanish, German, French, Hindi, and Chinese, tokenizing the space together with the answer letter consistently yields higher accuracy and lower ECE. Notably, even in Chinese -- a language not natively supported by Llama 3.1 \cite{meta2024llama3} -- we observe a substantial gain of over 4 accuracy points and a reduction of 5 ECE points. This confirms that the effect is not limited to English prompts and generalizes to multilingual settings, regardless of the model's native language capabilities.

\begin{table}
    \centering
    \small
    \begin{tabular}{lcccc}
        \toprule
        & \multicolumn{2}{c}{Accuracy (↑)} & \multicolumn{2}{c}{ECE (↓)} \\
        \cmidrule(lr){2-3} \cmidrule(lr){4-5}
        Language & ``\texttt{X}'' & ``\texttt{\textvisiblespace X}'' & ``\texttt{X}'' & ``\texttt{\textvisiblespace X}'' \\
        \midrule
        Spanish & 54.0 & \bfseries 56.5\textsuperscript{*} & 3.7 & \bfseries 1.1\textsuperscript{*} \\
        German & 52.9 & \bfseries 55.2\textsuperscript{*} & 3.4 & \bfseries 1.2\textsuperscript{*} \\
        French & 53.2 & \bfseries 56.3\textsuperscript{*} & 3.9 & \bfseries 1.1\textsuperscript{*} \\
        Hindi & 41.8 & \bfseries 45.5\textsuperscript{*} & 6.4 & \bfseries 2.1\textsuperscript{*} \\
        Chinese & 47.6 & \bfseries 51.9\textsuperscript{*} & 7.2 & \bfseries 2.1\textsuperscript{*} \\
        \bottomrule
    \end{tabular}
    \caption{Performance of Llama 3.1 8B on the MMLU benchmark in different languages.}
    \label{tab:languages}
\end{table}

\subsection{Performance across Datasets}
\label{app:datasets_results}

Below we provide the complete per-model, per-dataset results that complement the aggregated deltas shown in Figure~\ref{fig:datasets_deltas} of the main paper. We observe a consistent trend in favor of tokenizing the space together with the answer letter across datasets and model families. Even our largest evaluated model, Qwen~2.5~72B, shows a substantial gap of 11.7 accuracy points (on HellaSwag), indicating that larger models are also susceptible to such tokenization effects.

\begin{table*}[]
    \centering
    \small
    \setlength{\tabcolsep}{2pt}
    \begin{minipage}[t]{0.49\textwidth}
        \centering
        \begin{tabular}[t]{llcccc}
            \toprule
            & & \multicolumn{2}{c}{Acc ($\uparrow$)} & \multicolumn{2}{c}{ECE ($\downarrow$)} \\
            \cmidrule(lr){3-4} \cmidrule(lr){5-6}
            Model & Dataset & ``\texttt{X}'' & ``\texttt{\textvisiblespace X}'' & ``\texttt{X}'' & ``\texttt{\textvisiblespace X}'' \\
            \midrule
            \bfseries Llama 2 7B & ARC Challenge & 43.9 & \bfseries 46.0\textsuperscript{*} & 2.6 & \bfseries 2.2\phantom{\textsuperscript{*}} \\
            & ARC Easy & 57.6 & \bfseries 61.5\textsuperscript{*} & \bfseries 7.9 & 9.0\phantom{\textsuperscript{*}} \\
            & HellaSwag & 26.8 & \bfseries 29.4\textsuperscript{*} & 8.2 & \bfseries 3.7\textsuperscript{*} \\
            & OpenbookQA & 36.6 & \bfseries 39.4\textsuperscript{*} & 3.3 & \bfseries 2.9\phantom{\textsuperscript{*}} \\
            & TruthfulQA & 23.3 & \bfseries 24.7 & 4.9 & \bfseries 3.7\textsuperscript{*} \\
            \midrule
            \bfseries Llama 3.1 8B & ARC Challenge & 75.8 & \bfseries 78.8\textsuperscript{*} & 1.3 & 1.3\phantom{\textsuperscript{*}} \\
            & ARC Easy & 90.4 & \bfseries 91.6\textsuperscript{*} & 2.1 & \bfseries 1.9\phantom{\textsuperscript{*}} \\
            & HellaSwag & 46.8 & \bfseries 52.8\textsuperscript{*} & 4.9 & \bfseries 2.6\textsuperscript{*} \\
            & OpenbookQA & 73.2 & \bfseries 77.4\textsuperscript{*} & 2.4 & \bfseries 1.0\textsuperscript{*} \\
            & TruthfulQA & 43.9 & \bfseries 46.0\textsuperscript{*} & 7.8 & \bfseries 7.1\phantom{\textsuperscript{*}} \\
            \midrule
            \bfseries Llama 3.1 8B Inst & ARC Challenge & 82.1 & \bfseries 82.7\phantom{\textsuperscript{*}} & 1.8 & \bfseries 1.5\phantom{\textsuperscript{*}} \\
            & ARC Easy & 93.2 & \bfseries 93.5\phantom{\textsuperscript{*}} & 0.9 & \bfseries 0.2\textsuperscript{*} \\
            & HellaSwag & 51.4 & \bfseries 59.3\textsuperscript{*} & 6.4 & \bfseries 1.9\textsuperscript{*} \\
            & OpenbookQA & 81.4 & \bfseries 84.4\textsuperscript{*} & \bfseries 1.9 & 2.4\phantom{\textsuperscript{*}} \\
            & TruthfulQA & 56.1 & \bfseries 57.8\textsuperscript{*} & 8.7 & \bfseries 8.3\phantom{\textsuperscript{*}} \\
            \midrule
            \bfseries Llama 3.1 70B & ARC Challenge & 91.8 & \bfseries 91.9\phantom{\textsuperscript{*}} & 1.4 & \bfseries 1.2\phantom{\textsuperscript{*}} \\
            & ARC Easy & 97.2 & \bfseries 97.5\phantom{\textsuperscript{*}} & 1.5 & \bfseries 1.3\phantom{\textsuperscript{*}} \\
            & HellaSwag & 65.4 & \bfseries 68.5\textsuperscript{*} & \bfseries 2.7 & 3.0\phantom{\textsuperscript{*}} \\
            & OpenbookQA & 89.4 & \bfseries 90.8\phantom{\textsuperscript{*}} & 3.0 & \bfseries 2.7\phantom{\textsuperscript{*}} \\
            & TruthfulQA & 57.6 & \bfseries 67.4\textsuperscript{*} & 4.6 & \bfseries 2.1\textsuperscript{*} \\
            \midrule
            \bfseries Llama 3.1 70B Inst & ARC Challenge & 93.0 & \bfseries 94.2\textsuperscript{*} & 2.0 & \bfseries 0.3\textsuperscript{*} \\
            & ARC Easy & 97.7 & \bfseries 98.1\phantom{\textsuperscript{*}} & 2.0 & \bfseries 0.8\textsuperscript{*} \\
            & HellaSwag & 64.9 & \bfseries 68.1\textsuperscript{*} & 3.1 & \bfseries 1.2\textsuperscript{*} \\
            & OpenbookQA & 93.6 & \bfseries 94.2\phantom{\textsuperscript{*}} & 3.8 & \bfseries 2.1\textsuperscript{*} \\
            & TruthfulQA & 73.6 & \bfseries 75.8\textsuperscript{*} & \bfseries 2.6 & \bfseries 2.6\phantom{\textsuperscript{*}} \\
            \midrule
            \bfseries Gemma 3 4B & ARC Challenge & 70.6 & \bfseries 74.5\textsuperscript{*} & 5.2 & \bfseries 1.5\textsuperscript{*} \\
            & ARC Easy & 86.2 & \bfseries 88.6\textsuperscript{*} & 2.9 & \bfseries 0.8\textsuperscript{*} \\
            & HellaSwag & 37.3 & \bfseries 46.9\textsuperscript{*} & 8.9 & \bfseries 3.9\textsuperscript{*} \\
            & OpenbookQA & 58.8 & \bfseries 65.4\textsuperscript{*} & 6.0 & \bfseries 2.3\textsuperscript{*} \\
            & TruthfulQA & 31.0 & \bfseries 32.9\phantom{\textsuperscript{*}} & 15.0 & \bfseries 10.1\textsuperscript{*} \\
            \midrule
            \bfseries Gemma 3 4B Inst & ARC Challenge & 76.9 & \bfseries 77.0\phantom{\textsuperscript{*}} & 10.7 & \bfseries 10.6 \\
            & ARC Easy & 89.6 & \bfseries 89.9\phantom{\textsuperscript{*}} & \bfseries 4.7 & 4.8 \\
            & HellaSwag & 49.9 & \bfseries 50.4\phantom{\textsuperscript{*}} & 23.2 & \bfseries 22.9 \\
            & OpenbookQA & 74.0 & \bfseries 74.2\phantom{\textsuperscript{*}} & 12.3 & \bfseries 11.9 \\
            & TruthfulQA & 46.4 & \bfseries 47.0\phantom{\textsuperscript{*}} & 20.4 & \bfseries 19.8 \\
            \midrule
            \bfseries Gemma 3 12B & ARC Challenge & 89.0 & \bfseries 89.3\phantom{\textsuperscript{*}} & \bfseries 0.9 & 1.4\phantom{\textsuperscript{*}} \\
            & ARC Easy & 95.5 & \bfseries 96.0\phantom{\textsuperscript{*}} & 1.6 & \bfseries 0.6\textsuperscript{*} \\
            & HellaSwag & 50.8 & \bfseries 53.9\textsuperscript{*} & 4.7 & \bfseries 1.8\textsuperscript{*} \\
            & OpenbookQA & 83.4 & \bfseries 84.8\textsuperscript{*} & 4.7 & \bfseries 3.4\phantom{\textsuperscript{*}} \\
            & TruthfulQA & 51.4 & \bfseries 55.0\textsuperscript{*} & 6.8 & \bfseries 3.1\textsuperscript{*} \\
            \bottomrule
        \end{tabular}
    \end{minipage}%
    \hfill
    \begin{minipage}[t]{0.49\textwidth}
        \centering
        \begin{tabular}[t]{llcccc}
            \toprule
            & & \multicolumn{2}{c}{Acc ($\uparrow$)} & \multicolumn{2}{c}{ECE ($\downarrow$)} \\
            \cmidrule(lr){3-4} \cmidrule(lr){5-6}
            Model & Dataset & ``\texttt{X}'' & ``\texttt{\textvisiblespace X}'' & ``\texttt{X}'' & ``\texttt{\textvisiblespace X}'' \\
            \midrule
            \bfseries Mistral 7B v0.3 & ARC Challenge & 75.6 & \bfseries 76.7 & 1.9 & \bfseries 1.8\phantom{\textsuperscript{*}} \\
            & ARC Easy & 88.5 & \bfseries 88.7 & 2.1 & \bfseries 2.0\phantom{\textsuperscript{*}} \\
            & HellaSwag & 46.7 & \bfseries 47.8 & \bfseries 2.6 & 3.3\phantom{\textsuperscript{*}} \\
            & OpenbookQA & 72.4 & \bfseries 73.4 & 3.0 & \bfseries 2.9\phantom{\textsuperscript{*}} \\
            & TruthfulQA & 45.0 & \bfseries 45.4 & 5.3 & \bfseries 3.9\textsuperscript{*} \\
            \midrule
            \bfseries Mistral 7B Inst v0.3 & ARC Challenge & 78.4 & \bfseries 78.6\phantom{\textsuperscript{*}} & 6.3 & \bfseries 5.7 \\
            & ARC Easy & 88.3 & \bfseries 88.9\phantom{\textsuperscript{*}} & 3.1 & \bfseries 2.7 \\
            & HellaSwag & 60.0 & \bfseries 61.1\phantom{\textsuperscript{*}} & 9.6 & \bfseries 8.9 \\
            & OpenbookQA & 77.2 & \bfseries 77.6\phantom{\textsuperscript{*}} & 5.9 & \bfseries 5.0 \\
            & TruthfulQA & 48.6 & \bfseries 49.9\textsuperscript{*} & 13.1 & \bfseries 13.0 \\
            \midrule
            \bfseries Mistral Small 24B & ARC Challenge & 92.2 & \bfseries 92.7 & 1.4 & \bfseries 1.2\phantom{\textsuperscript{*}} \\
            & ARC Easy & 97.7 & \bfseries 97.8 & 1.9 & \bfseries 1.7\phantom{\textsuperscript{*}} \\
            & HellaSwag & 55.4 & \bfseries 56.0 & 4.1 & \bfseries 3.9\phantom{\textsuperscript{*}} \\
            & OpenbookQA & 86.2 & \bfseries 86.4 & \bfseries 3.3 & 4.4\phantom{\textsuperscript{*}} \\
            & TruthfulQA & 67.3 & \bfseries 68.2 & \bfseries 1.5 & \bfseries 1.5\phantom{\textsuperscript{*}} \\
            \midrule
            \bfseries Qwen 2.5 7B & ARC Challenge & 88.3 & \bfseries 88.7\phantom{\textsuperscript{*}} & 1.4 & \bfseries 0.4\textsuperscript{*} \\
            & ARC Easy & 95.4 & \bfseries 96.7\textsuperscript{*} & 1.6 & \bfseries 0.9\textsuperscript{*} \\
            & HellaSwag & 59.6 & \bfseries 63.0\textsuperscript{*} & 1.2 & \bfseries 0.6\textsuperscript{*} \\
            & OpenbookQA & 86.2 & \bfseries 89.0\textsuperscript{*} & 2.5 & \bfseries 1.7\textsuperscript{*} \\
            & TruthfulQA & 60.8 & \bfseries 63.3\textsuperscript{*} & \bfseries 2.4 & 3.2\phantom{\textsuperscript{*}} \\
            \midrule
            \bfseries Qwen 2.5 72B & ARC Challenge & 95.6 & \bfseries 95.8\phantom{\textsuperscript{*}} & 2.3 & \bfseries 0.4\textsuperscript{*} \\
            & ARC Easy & 98.5 & \bfseries 98.8\phantom{\textsuperscript{*}} & 1.8 & \bfseries 0.4\textsuperscript{*} \\
            & HellaSwag & 76.7 & \bfseries 78.0\textsuperscript{*} & 3.7 & \bfseries 2.7\textsuperscript{*} \\
            & OpenbookQA & 95.8 & \bfseries 96.4\phantom{\textsuperscript{*}} & 4.5 & \bfseries 2.0\textsuperscript{*} \\
            & TruthfulQA & 63.3 & \bfseries 75.0\textsuperscript{*} & 3.2 & \bfseries 1.9\textsuperscript{*} \\
            \midrule
            \bfseries Qwen 3 8B & ARC Challenge & 90.5 & \bfseries 92.1\textsuperscript{*} & 1.1 & \bfseries 1.0\phantom{\textsuperscript{*}} \\
            & ARC Easy & 97.3 & \bfseries 97.6\phantom{\textsuperscript{*}} & 1.5 & \bfseries 1.3\phantom{\textsuperscript{*}} \\
            & HellaSwag & 63.4 & \bfseries 68.0\textsuperscript{*} & 3.5 & \bfseries 2.1\textsuperscript{*} \\
            & OpenbookQA & 85.6 & \bfseries 85.6\phantom{\textsuperscript{*}} & \bfseries 2.3 & 2.8\phantom{\textsuperscript{*}} \\
            & TruthfulQA & 60.2 & \bfseries 64.9\textsuperscript{*} & 3.2 & \bfseries 3.0\phantom{\textsuperscript{*}} \\
            \midrule
            \bfseries GPT Neo 2.7B & ARC Challenge & 23.8 & \bfseries 24.7\phantom{\textsuperscript{*}} & 9.4 & \bfseries 3.0\textsuperscript{*} \\
            & ARC Easy & 24.6 & \bfseries 26.7\textsuperscript{*} & 10.0 & \bfseries 2.8\textsuperscript{*} \\
            & HellaSwag & 26.2 & \bfseries 27.0\textsuperscript{*} & 14.1 & \bfseries 4.8\textsuperscript{*} \\
            & OpenbookQA & 24.8 & \bfseries 26.2\textsuperscript{*} & 12.3 & \bfseries 2.9\textsuperscript{*} \\
            & TruthfulQA & 22.4 & \bfseries 22.8\phantom{\textsuperscript{*}} & 7.5 & \bfseries 4.4\textsuperscript{*} \\
            \bottomrule
        \end{tabular}
    \end{minipage}
    \caption{Full results of all models on all datasets tokenizing the space before (Letter token; ``\texttt{X}'') or together with the letter (Space--Letter token; ``\texttt{\textvisiblespace X}''). \textsuperscript{*}~means significantly better (higher for accuracy; lower for ECE). Top-performing tokenization strategy for each model is bolded.}
    \label{tab:full_results}
\end{table*}

\end{document}